\pdfoutput=1

\documentclass[11pt]{article}

\usepackage[preprint]{acl}

\usepackage{times}
\usepackage{latexsym}

\usepackage[T1]{fontenc}

\usepackage[utf8]{inputenc}

\usepackage{microtype}
\usepackage{booktabs}
\usepackage{inconsolata}

\usepackage{graphicx}

\usepackage{algorithm}
\usepackage{algpseudocode}
\usepackage{amsmath}
\usepackage{amssymb}
\usepackage{cleveref}
\usepackage{authblk}

\usepackage[most]{tcolorbox}
\usepackage{enumitem}

\usepackage{xspace}
\newcommand{\ours}{\texttt{Sub-CP}\xspace}

\newcommand{\sA}{\mathcal{A}}
\newcommand{\sB}{\mathcal{B}}
\newcommand{\sV}{\mathcal{V}}

\title{Submodular Context Partitioning for In-Context Learning}

\author{
  Shaoyi Zheng\textsuperscript{1} \quad
  Canyu Zhang\textsuperscript{1} \quad
  Lilly Kumari\textsuperscript{2} \quad
  Tianyi Zhou\textsuperscript{3} \quad
  Shengjie Wang\textsuperscript{1} \\
  \textsuperscript{1}New York University \\
  \textsuperscript{2}University of Washington, Seattle \\
  \textsuperscript{3}University of Maryland, College Park \\
}

\begin{document}
\maketitle
\begin{abstract}
In-context learning (ICL) enables efficient few-shot learning in large language models (LLMs) without training, but suffers from the quadratic input complexity of transformers, limiting the maximum number of exemplars.
While various efficient ICL approaches partition the context into blocks to process (e.g., ensembling, compression, cross-attention), they often ignore the information redundancy or under-representation caused by different partition strategies, leading to suboptimal performance. 
To tackle this problem, we propose \textbf{\ours}, a block-aware context selection framework that leverages submodular objectives to control block diversity. \ours supports a flexible spectrum of selection strategies, allowing each block to range from globally diverse to locally coherent. This allows fine-grained control over semantic structure while enabling precomputation.
Extensive experiments across diverse tasks on multiple datasets show that \ours consistently improves performance across model scales.
\end{abstract}

\section{Introduction}
Large Language Models (LLMs) such as GPT-4o~\cite{openai2024gpt4o}, DeepSeek-V2~\cite{deepseek2024v2}, and LLaMA~\cite{touvron2023llama, touvron2023llama2} exhibit strong few-shot generalization via \textbf{in-context learning (ICL)}~\cite{brown2020language}, where models address new queries/tasks by learning from input exemplars/demonstrations during inference. However, the quadratic complexity of transformer-based inference~\cite{vaswani2017attention} limits ICL scalability to many exemplars due to the long context.

Numerous methods have been proposed to reduce context length while preserving task-relevant information to address this. These methods often partition the context into blocks and perform block-wise operations to reduce cost, e.g., ensembling~\cite{zhang2023block}, compression~\cite{li2023contextcompressor}, and cross-attention~\cite{fu2024giraffe}. While effective in saving computation, these methods often rely on random or uniform partitioning of examples, overlooking the structure of information within and across blocks. As a result, they may select suboptimal context blocks due to intra-block redundancy or under-representation of critical exemplars, degrading the ICL performance. 

What partition strategy can produce the best blocks of exemplars for efficient ICL? We propose \textbf{Submodular Context Partitioning (\ours)}, a block-aware context selection framework that leverages submodular objectives to control the degree of diversity within each block. Submodular function is a rich class of functions that describe the diversity of a given set of inputs. By adjusting the submodular objective used to evaluate information coverage, \ours creates a spectrum that spans two ends of block information: global diversity and local coherence. Specifically, \ours uses four strategies: \textbf{Global Diverse} maximizes overall diversity by encouraging each block to cover distinct and representative portions of the entire dataset; \textbf{Global-Local Diverse} balances between global coverage and inter-block complementarity, ensuring blocks are both individually informative and mutually distinctive. \textbf{Local Diverse} promotes intra-block diversity by selecting most diverse examples within cluster, enhancing the expressiveness of local contents. \textbf{Local Coherent} prioritizes semantic consistency within each block by selecting contexts concentrated around cluster centroids. These design choices enable \ours to have flexible control over the semantic structure of each block as well as the dependency across blocks. They are complementary to existing efficient ICL methods and can be applied as a pre-processing step in any of them.

\begin{figure*}[t!]
  \centering
  \includegraphics[width=0.92\textwidth,trim=2cm 2cm 3cm 2.2cm,clip]{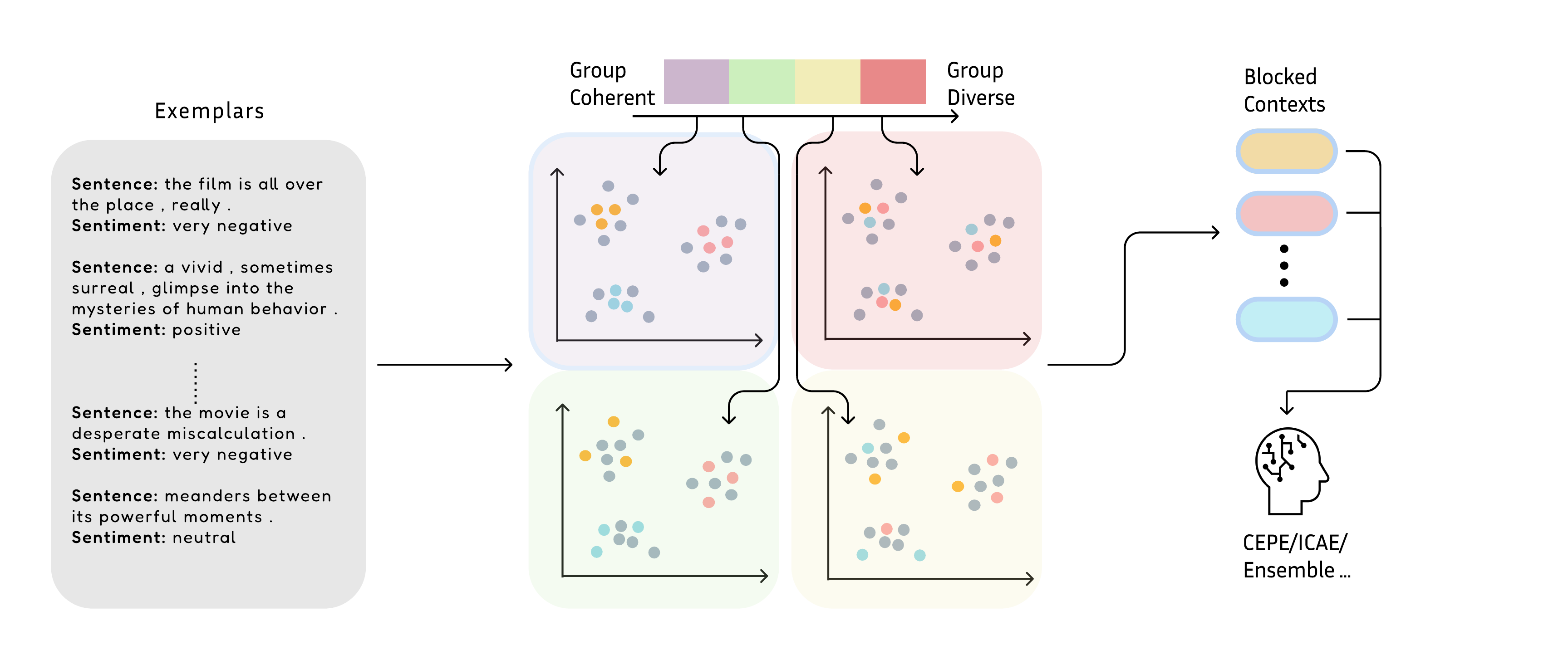}
  \caption{The \ours overview. \ours selects and partitions from the whole dataset and then distributes them into blocks in four ways, ranging from global diverse to local coherent.}
  \label{fig:pipeline}
\end{figure*}

We evaluate \ours on three efficient ICL frameworks: In-context Autoencoder (ICAE), Demonstration Ensemble (DENSE), and Context Expansion with Parallel Encoding (CEPE), using several classification benchmarks. \ours yields varying degrees of performance improvement, demonstrating its flexibility and potential in diverse ICL settings. These results underscore the importance of structure-aware context construction in enabling scalable and effective ICL under limited inference budgets. \looseness=-1

\section{Method}
\ours is a general framework pluggable with any submodular function to model the diversity of blocks.
Particularly, we use the facility location function, which relies on a similarity matrix $S \in \mathbb{R}^{N \times N}$ 
computed on example embeddings. The facility location function is $f(G) = \sum_j \max_{i \in G} S_{i,j}$, which quantifies the information coverage of selected indices $G$ over the dataset.




\paragraph{Global Diverse} encourages each block to capture a diverse summary of the dataset. At each step, we add context that provides the highest marginal gain (or the most diverse one given previous selections) to that block with the lowest submodular score. This iterative process ensures that even the least diverse block accumulates varied information. As a result, with globally diverse contexts, all blocks achieve nearly full information coverage, and each serves as a compact summary of the whole dataset distribution. \looseness=-1

\paragraph{Global-Local Diverse} extends the Global Diverse strategy to mitigate information redundancy across blocks. While Global Diverse encourages each block to independently capture a representative summary of the full dataset, it may result in significant overlap between blocks. In contrast, Global-Local Diverse explicitly promotes inter-block complementarity by selecting contexts that are informative to the current block while remaining distinct from previously selected blocks. Formally, we redefine the submodular score as \( f(B \mid A) + f(B) \), where \( A \) denotes the current block and \( B \) denotes the union of all previously selected blocks. Compared to Local Diverse, this method still selects globally, but achieves soft locality through inter-block separation, thereby implicitly capturing localized information while maintaining global coverage. \looseness=-1

\paragraph{Local Diverse}
In contrast to Global Diverse and Global-Local Diverse, which aim for each block to summarize the entire dataset, Local Diverse seeks to preserve the internal diversity of localized data regions. Specifically, we first partition the dataset into semantically coherent clusters and then apply a Global Diverse-style selection procedure within each cluster to select context with the most information gain within this cluster, ensuring intra-cluster diversity. This results in locally diverse blocks while remaining complementary across clusters to maintain global coverage. 

\paragraph{Local Coherent}
While previous strategies focus on enhancing diversity globally or locally, Local Coherent aims to construct blocks that are semantically consistent and aligned within a specific subregion of the dataset. Rather than selecting examples that span a wide range of variation, Local Coherent promotes contextual coherence by selecting examples that are similar to the core of each semantic cluster. Concretely, we first apply Global Diverse-style selection to identify diverse centroids as representatives of distinct clusters. Then, for each centroid, we select contexts that are most similar to it, forming a block around a coherent semantic theme. This encourages each block to maintain internal consistency and topic alignment, which can be beneficial for tasks that rely on local patterns or subtle variations within a concept space.

As shown in Fig.~\ref{fig:pipeline}, we propose four algorithms that leverage submodular functions to control the diversity and structure of context blocks. These methods form a spectrum ranging from global diversity to local coherence, enabling flexible control over block content and inter-block relationships to suit different in-context learning needs. Detailed algorithmic procedures are provided in the Appendix.


\section{Experiment Result}
\paragraph{Dataset} We tested 5 classification datasets: SST-2~\cite{socher2013recursive}, SST-5~\cite{socher2013recursive}, MR~\cite{pang2005seeing}, TREC~\cite{li2002learning}, and AG News~\cite{zhang2015character} 
\paragraph{Experiment}
Our goal is to evaluate whether submodular-based context partitioning improves in-context learning performance against uniform and random sampling across multiple ICL frameworks. We fix the number of shots to 40 for all experiments and report normalized accuracy, averaged over three random seeds. \looseness=-1

\subsection{Demonstration Ensembling (DENSE)}
\paragraph{Baseline}
Instead of aggregating all demonstrations into a single prompt, DENSE partitions the demonstrations into multiple subsets, queries the language model separately for each subset, and aggregates the resulting predictions through different ensembling strategies. We consider three ensembling strategies: \textbf{PoE} multiplies prediction probabilities to reinforce consensus; \textbf{MoE} averages predictions for balanced integration; and \textbf{Max} selects the most confident prediction across subsets. We reimplement DENSE using Mistral-7B-v2, replacing the original LLaMA2-7B for a fairer and more modern comparison across all experiments.
We apply \ours and use each block as one ensemble instance for testing our performance. \looseness=-1
\begin{table}[h]
\centering
\Large
\resizebox{\linewidth}{!}{%
\begin{tabular}{l|cccc|c}
\toprule
\textbf{Ensemble} & \textbf{Local Co.} & \textbf{Global Div.} & \textbf{Global-Local Div.} & \textbf{Local Div.} & \textbf{Baseline} \\
\midrule
\multicolumn{6}{l}{\textit{MoE}} \\
AG News & \textbf{65.83} & 65.25 & 64.59 & 65.82 & 63.13 \\
SST-2  & \textbf{94.50} & 92.78 & 92.78 & 93.12 & 94.27 \\
SST-5  & 43.05 & 44.23 & 44.32 & \textbf{46.05} & 36.33 \\
MR      & 89.59 & \textbf{90.90} & 89.40 & 90.15 & 91.46 \\
TREC    & 56.86 & \textbf{57.80} & 56.40 & 57.60 & 42.20 \\
\midrule
\multicolumn{6}{l}{\textit{PoE}} \\
AG News & 61.07 & 62.51 & 60.46 & \textbf{62.92} & 55.03 \\
SST-2  & \textbf{92.66} & 90.48 & 91.40 & 91.17 & 91.74 \\
SST-5  & 43.05 & 44.32 & 44.60 & \textbf{45.96} & 37.33 \\
MR      & 80.68 & 85.37 & 83.68 & \textbf{86.12} & 88.27 \\
TREC    & 64.00 & 64.20 & 64.40 & \textbf{67.60} & 38.40 \\
\midrule
\multicolumn{6}{l}{\textit{Max}} \\
AG News & 65.01 & 64.74 & 63.68 & \textbf{65.29} & 55.66 \\
SST-2  & \textbf{94.61} & 92.55 & 93.12 & 93.12 & 94.38 \\
SST-5  & 42.05 & 43.87 & 43.75 & \textbf{46.50} & 35.69 \\
MR      & 86.49 & 89.12 & 87.99 & \textbf{88.84} & 91.46 \\
TREC    & 59.80 & \textbf{64.00} & 63.60 & 62.20 & 40.20 \\
\midrule
\textbf{AVG} & 69.35 & 70.14 & 69.62 & \textbf{70.83} & 63.70 \\
\bottomrule
\end{tabular}
}
\caption{Performance comparison (\%) of different context selection methods under three ensemble strategies. Bold indicates the best score in each row.}
\label{tab:ensemble_results}
\end{table}

\paragraph{Qualitative Comparison}
From Tab.~\ref{tab:ensemble_results}, we observe that \ours series consistently performs well across all three ensembling variants. On high-accuracy datasets such as SST-2 and MR, our methods maintain performance that is comparable to or even exceeds that of the baseline, demonstrating that they do not compromise accuracy in easy settings. More importantly, for more challenging datasets like SST-5 and TREC, we observe substantial improvements across the board. For instance, Local Diverse under the PoE ensemble achieves a remarkable +29.2\% absolute gain over the baseline on TREC, while similar trends hold for other ensemble methods as well. These consistent gains highlight the robustness and generalizability of our block-partitioned context selection approach.

\subsection{In-context Autoencoder (ICAE)}
\paragraph{Baseline} ICAE is a learnable compression framework that reduces long contexts into compact memory slots fed to an LLM. It uses a lightweight LoRA-trained encoder and a frozen decoder (the LLM), pretrained with autoencoding and LM objectives, then fine-tuned on instructions. In ICAE, we separately compress each blocks and then fed to LLM and record the performance.

\paragraph{Qualitative Comparison}
Tab.~\ref{tab:icae_results} shows ICAE performance across five classification benchmarks using different block selection strategies. Global-Local Diverse and Local Diverse outperform others, achieving the highest average accuracy of 66.26\% and 65.75\% respectively, substantially surpassing the random baseline (60.92\%). Notably, Global-Local Diverse excels on TREC, while Local Diverse dominates on AG News, SST-5 and MR, highlighting its ability to balance intra-cluster diversity with global representation. \looseness=-1

\begin{table}[h]
\centering
\scriptsize
\resizebox{\linewidth}{!}{%
\begin{tabular}{l|cccc|c}
\toprule
\textbf{ICAE Dataset} & \textbf{Local Co.} & \textbf{Global Div.} & \textbf{ Global-Local Div.} & \textbf{Local Div.} & \textbf{Baseline} \\
\midrule
AG News       & 71.93 & 62.45 & 74.42 & \textbf{78.33} & 53.78 \\
SST-2        & 90.13 & \textbf{92.88} & 91.16 & 91.39 & 88.65 \\
SST-5        & 26.60 & 26.79 & 21.90 & \textbf{30.81} & 31.36 \\
TREC   & 39.20 & 43.60 & \textbf{59.00} & 37.80 & 45.80 \\
MR            & 86.11 & 86.30 & 84.80 & \textbf{90.43} & 84.99 \\
\midrule
\textbf{AVG} & 62.79 & 62.40 & 66.26 & \textbf{65.75} & 60.92 \\
\bottomrule
\end{tabular}
}
\caption{ICAE performance comparison (\%) across different block selection methods and a random baseline. Bold indicates the best performance per dataset.}
\label{tab:icae_results}
\end{table}

\subsection{Context Expansion with Parallel Encoding (CEPE)}
\paragraph{Baseline}
CEPE is a lightweight framework for extending the context window of decoder-only large language models (LLMs). It encodes context either jointly (Concat) or independently (Separate), and integrates it into the decoder via a cross-attention mechanism at each layer. This design allows the decoder to incorporate information from the encoder’s representations while significantly reducing computational complexity.

\begin{table}[h]
\centering
\Large
\resizebox{\linewidth}{!}{%
\begin{tabular}{l|cccc|cc}
\toprule
\textbf{CEPE Dataset} & \textbf{Local Co.} & \textbf{Global Div.} & \textbf{ Global-Local Div.} & \textbf{Local Div.} & \textbf{Concat} & \textbf{Separate} \\
\midrule
AG News       & 61.33 & \textbf{70.17} & 67.17 & 62.01 & 63.02 & 66.45 \\
SST-2        & 85.71 & \textbf{91.11} & 66.97 & 82.14 & 78.57 & 86.90 \\
SST-5        & \textbf{42.59} & 41.24 & 38.41 & 40.51 & 40.59 & 42.33 \\
TREC   & 43.00 & 54.00 & \textbf{60.00} & 49.20 & 54.40 & 41.19 \\
MR            & 88.18 & 88.08 & 88.64 & \textbf{88.84} & 92.16 & 93.14 \\
\midrule
\textbf{AVG} & 64.16 & \textbf{68.92} & 64.24 & 64.54 & 65.75 & 66.00 \\
\bottomrule
\end{tabular}
}
\caption{CEPE performance comparison (\%) across different block Concat and Separate. Bold indicates the best performance for each dataset.}
\label{tab:cepe_results}
\end{table}
\vspace{-10pt}

\paragraph{Qualitative Comparison}
From Tab.~\ref{tab:cepe_results}, across all datasets under the CEPE, \textbf{Global Diverse} consistently achieves the highest average performance, outperforming random baselines and other structured selection methods. While cluster-based variants like \textbf{Local Diverse} and \textbf{Global-Local Diverse} show strengths on specific datasets, Global Diverse provides the most robust improvement overall, highlighting the effectiveness of globally diverse selection in enhancing model performance. 

\section{Related Work}
Existing ICL methods include \textbf{learnable compression} approaches (e.g., AutoCompressor~\cite{wu2023autocompressor}, GIST~\cite{li2023gist}) that require training, and \textbf{retrieval-based} methods (e.g., AutoICL~\cite{xu2023autoicl}, ICCL~\cite{zhang2022iccl}) that rely on query similarity, resulting in limited efficiency and generality. In contrast, \textbf{\ours} requires diverse and coherent context blocks through query-independent submodular function.

\section{Discussion}
\paragraph{Analysis on CEPE}
While \ours consistently improves performance across ICL frameworks, its gains under CEPE are relatively modest. Further analysis reveals that CEPE performs best in the \textit{separate} setting, where each context is encoded independently, and degrades notably when using larger block sizes, indicating limited sensitivity to block-level embedding quality. This behavior may stem from CEPE’s use of cross-attention without positional encodings: in the separate setting, block order has minimal impact, whereas concatenated blocks introduce order-sensitive variance. As a result, CEPE cannot fully exploit the structured representations provided by our block selection strategy. \looseness=-1

\paragraph{Analysis on query-dependent}
Currently, our method is query-independent and operates at the dataset level, allowing for precomputation. While query-aware selection can improve relevance, relying solely on query similarity may result in under-representative contexts. A promising direction is to combine both strategies—precomputing diverse context blocks offline and augmenting them with a few query-relevant examples at inference time—to strike a balance between coverage and specificity.

\paragraph{Analysis on Label and Size Balance}
Since our method operates at the embedding level, selecting diverse or coherent contexts implicitly encourages a certain degree of label and size balance across blocks, assuming label information is encoded in the embeddings. However, we observe that such implicit control may be insufficient, particularly on challenging datasets. By explicitly enforcing label and sample count balance within each block, we achieve notable performance gains on underperforming datasets such as SST-5 under the CEPE framework, around 4\% in Global Diverse. This introduces a trade-off: while explicit balancing can improve robustness, it may also restrict the flexibility of submodular selection, potentially limiting performance in cases where semantic diversity is more beneficial. \looseness=-1

\section{Conclusion}
We propose \textbf{\ours}, a submodular-based block-aware context selection in ICL. By explicitly controlling block content to range from global diversity to local coherence, \ours effectively improves performance on different ICL frameworks. It shows that \ours consistently delivers notable performance gains on most models.

\newpage
\section{Limitations}
While \textbf{\ours} consistently improves performance across various ICL frameworks, several limitations remain. First, under the CEPE setting, the gains are less pronounced. We find that CEPE benefits more from independently encoded (separate) contexts, where block ordering has minimal effect due to the lack of positional encoding in cross-attention. This limits the advantage of \ours’s block-level optimization.

Second, \ours is currently query-independent, allowing efficient offline computation but potentially missing query-specific relevance. A hybrid design that combines offline selection with lightweight query-aware adjustment may offer better trade-offs.

Lastly, as \ours selects contexts based on semantic similarity in embedding space, it implicitly assumes balanced label and sample distribution. We observe that explicitly enforcing label and number balance can improve performance on certain datasets (e.g., SST-5), though this may reduce the flexibility of submodular optimization in some scenarios.

\bibliography{custom}

\begin{thebibliography}{19}
\providecommand{\natexlab}[1]{#1}

\bibitem[{AI(2024)}]{deepseek2024v2}
DeepSeek AI. 2024.
\newblock Deepseek-v2: Towards deeper language understanding via long-context modeling.
\newblock \url{https://github.com/deepseek-ai/deepseek-v2}.
\newblock Accessed May 2024.

\bibitem[{Brown et~al.(2020)Brown, Mann, Ryder, Subbiah, Kaplan, Dhariwal, Neelakantan, Shyam, Sastry, Askell et~al.}]{brown2020language}
Tom~B Brown, Benjamin Mann, Nick Ryder, Melanie Subbiah, Jared Kaplan, Prafulla Dhariwal, Arvind Neelakantan, Pranav Shyam, Girish Sastry, Amanda Askell, and 1 others. 2020.
\newblock Language models are few-shot learners.
\newblock In \emph{Advances in Neural Information Processing Systems}.

\bibitem[{Fu et~al.(2024)Fu, Ge, Li, and Liu}]{fu2024giraffe}
Yizhou Fu, Yuxuan Ge, Weizhe Li, and Yisen Liu. 2024.
\newblock Giraffe: Long context understanding in llms via representational fusion.
\newblock \emph{arXiv preprint arXiv:2402.16617}.

\bibitem[{Fujishige(2005)}]{fujishige2005submodular}
Satoru Fujishige. 2005.
\newblock \emph{Submodular functions and optimization}, volume~58.
\newblock Elsevier.

\bibitem[{Li et~al.(2023{\natexlab{a}})Li, Wang, Ge, Liu et~al.}]{li2023contextcompressor}
Weizhe Li, Yutong Wang, Yuxuan Ge, Yisen Liu, and 1 others. 2023{\natexlab{a}}.
\newblock Context compressor: Context optimization via diffusion model for in-context learning.
\newblock \emph{arXiv preprint arXiv:2307.06945}.

\bibitem[{Li et~al.(2023{\natexlab{b}})Li, Zhang, Chen, Liang, He, Chen, and Ma}]{li2023gist}
Xiang Li, Zichao Zhang, Xuezhe Chen, Yingyu Liang, Xilun He, Yichong Chen, and Tengyu Ma. 2023{\natexlab{b}}.
\newblock Gist: Interpretable and generalizable in-context learning with selection and transformation.
\newblock In \emph{ICLR}.

\bibitem[{Li and Roth(2002)}]{li2002learning}
Xin Li and Dan Roth. 2002.
\newblock Learning question classification.
\newblock In \emph{COLING}, pages 556--562.

\bibitem[{Nemhauser et~al.(1978)Nemhauser, Wolsey, and Fisher}]{nemhauser1978analysis}
George~L Nemhauser, Laurence~A Wolsey, and Marshall~L Fisher. 1978.
\newblock An analysis of approximations for maximizing submodular set functions—i.
\newblock \emph{Mathematical programming}, 14:265--294.

\bibitem[{OpenAI(2024)}]{openai2024gpt4o}
OpenAI. 2024.
\newblock Gpt-4o technical report.
\newblock \url{https://openai.com/index/gpt-4o}.
\newblock Accessed May 2024.

\bibitem[{Pang and Lee(2005)}]{pang2005seeing}
Bo~Pang and Lillian Lee. 2005.
\newblock Seeing stars: Exploiting class relationships for sentiment categorization with respect to rating scales.
\newblock In \emph{ACL}, pages 115--124.

\bibitem[{Socher et~al.(2013)Socher, Perelygin, Wu, Chuang, Manning, Ng, and Potts}]{socher2013recursive}
Richard Socher, Alex Perelygin, Jean Wu, Jason Chuang, Christopher~D Manning, Andrew~Y Ng, and Christopher Potts. 2013.
\newblock Recursive deep models for semantic compositionality over a sentiment treebank.
\newblock In \emph{EMNLP}, pages 1631--1642.

\bibitem[{Touvron et~al.(2023{\natexlab{a}})Touvron, Lavril, Izacard, Martinet, Lachaux, Lacroix, Rozi{\`e}re, Goyal, Hambro, Azhar et~al.}]{touvron2023llama}
Hugo Touvron, Thibaut Lavril, Gautier Izacard, Xavier Martinet, Marie-Anne Lachaux, Timothee Lacroix, Baptiste Rozi{\`e}re, Naman Goyal, Eric Hambro, Faisal Azhar, and 1 others. 2023{\natexlab{a}}.
\newblock Llama: Open and efficient foundation language models.
\newblock \emph{arXiv preprint arXiv:2302.13971}.

\bibitem[{Touvron et~al.(2023{\natexlab{b}})Touvron, Martin, Stone, Albert, Almahairi, Babaei, Bashlykov, Batra, Bhargava, Bhosale et~al.}]{touvron2023llama2}
Hugo Touvron, Louis Martin, Kevin Stone, Peter Albert, Amjad Almahairi, Yasmine Babaei, Sergey Bashlykov, Trishul Batra, Prajjwal Bhargava, Shruti Bhosale, and 1 others. 2023{\natexlab{b}}.
\newblock Llama 2: Open foundation and fine-tuned chat models.
\newblock \emph{arXiv preprint arXiv:2307.09288}.

\bibitem[{Vaswani et~al.(2017)Vaswani, Shazeer, Parmar, Uszkoreit, Jones, Gomez, Kaiser, and Polosukhin}]{vaswani2017attention}
Ashish Vaswani, Noam Shazeer, Niki Parmar, Jakob Uszkoreit, Llion Jones, Aidan~N Gomez, {\L}ukasz Kaiser, and Illia Polosukhin. 2017.
\newblock Attention is all you need.
\newblock In \emph{Advances in Neural Information Processing Systems (NeurIPS)}.

\bibitem[{Wu et~al.(2023)Wu, Xu, Liu, Wang, Liu, and Sun}]{wu2023autocompressor}
Ziyang Wu, Canwen Xu, Xiao Liu, Yanlin Wang, Zhiyuan Liu, and Maosong Sun. 2023.
\newblock Autocompressor: Training-free context compression for in-context learning.
\newblock In \emph{EMNLP}.

\bibitem[{Xu et~al.(2023)Xu, Wu, Liu, Wang, and Sun}]{xu2023autoicl}
Canwen Xu, Ziyang Wu, Xiao Liu, Yanlin Wang, and Maosong Sun. 2023.
\newblock Autoicl: Automatically constructing demonstrations for in-context learning.
\newblock In \emph{ACL}.

\bibitem[{Zhang et~al.(2023)Zhang, Lin, Zhu, Ding, Zhang, Hou, Qian, Sun, and Yang}]{zhang2023block}
Jiawei Zhang, Zhengyan Lin, Zhengxiao Zhu, Ning Ding, Yuxian Zhang, Yujie Hou, Yuxiao Qian, Maosong Sun, and Zhilin Yang. 2023.
\newblock Block icl: Context optimization via blockwise learning for in-context learning.
\newblock \emph{arXiv preprint arXiv:2308.08780}.

\bibitem[{Zhang et~al.(2015)Zhang, Zhao, and LeCun}]{zhang2015character}
Xiang Zhang, Junbo Zhao, and Yann LeCun. 2015.
\newblock Character-level convolutional networks for text classification.
\newblock In \emph{NeurIPS}, pages 649--657.

\bibitem[{Zhang et~al.(2022)Zhang, Deng, Liu, and Zhou}]{zhang2022iccl}
Zhengbao Zhang, Yuxuan Deng, Zhiyuan Liu, and Jie Zhou. 2022.
\newblock Iccl: Instance-wise control for compositional generalization in prompt learning.
\newblock In \emph{NeurIPS}.

\end{thebibliography}

\clearpage
\appendix

\section{Submodularity}
\label{sec:appendix}

\subsection{Submodularity}

A submodular function~\cite{fujishige2005submodular} is a set function \( f: 2^{\sV} \rightarrow \mathbb{R} \) defined over subsets of a ground set \( \sV \). It satisfies the \textit{diminishing returns} property, which states that the marginal gain of adding a new element \( v \) to a set decreases as the set grows. Formally, for any subsets \( \sA \subseteq \sB \subseteq \sV \) and any element \( v \in \sV \setminus \sB \),
\[
f(v \mid \sA) \geq f(v \mid \sB),
\]
where the marginal gain is defined as
\[
f(v \mid \sA) := f(\sA \cup \{v\}) - f(\sA).
\]

This property makes submodular functions particularly suitable for modeling coverage and diversity in subset selection problems. A canonical formulation is submodular maximization under a cardinality constraint:
\[
\max_{\sA \subseteq \sV} \; f(\sA) \quad \text{s.t.} \; |\sA| \leq k.
\]

An intuitive and widely-used method for this problem is the greedy algorithm, which achieves a provable \((1 - 1/e)\)-approximation of the optimal solution~\cite{nemhauser1978analysis}. The algorithm starts with an empty set and iteratively adds the element with the largest marginal gain until the size constraint is met, as shown in Alg.~\ref{alg:greedy}.

\begin{algorithm}[h]
\caption{Greedy Submodular Maximization}
\label{alg:greedy}
\begin{algorithmic}[1]
\Require Ground set \( \mathcal{V} \), submodular function \( f: 2^{\mathcal{V}} \rightarrow \mathbb{R} \), budget \( k \)
\Ensure Selected subset \( \mathcal{A} \subseteq \mathcal{V} \) with \( |\mathcal{A}| \leq k \)

\State Initialize \( \mathcal{A} \gets \emptyset \)
\For{$i = 1$ to $k$}
    \State \( v^\star \gets \arg\max_{v \in \mathcal{V} \setminus \mathcal{A}} f(v \mid \mathcal{A}) \)
    \State \( \mathcal{A} \gets \mathcal{A} \cup \{v^\star\} \)
\EndFor
\State \Return \( \mathcal{A} \)
\end{algorithmic}
\end{algorithm}

\subsection{Objectives for Partitioning}

Let $\Pi(V, \mathcal{C})$ denote all possible partitionings of the groundset $V$ satisfying some constraint $\mathcal{C}$, e.g., every block has a limited size, label balance, or is inside a given cluster. $\pi \in \Pi(V, \mathcal{C})$ is one partitioning we optimize for. We propose to use the following four submodular partitioning objectives to generate ICL example blocks with various diversity and coherence.

\textbf{Global Diverse} selects diverse information from the entire dataset for every block, ensuring that each block serves as a compact summary of the full dataset.
$$\max_{\pi \in \Pi(V, \mathcal{C})} \min_{A_i \in \pi} f(A_i)$$

\textbf{Global-Local Diverse} ensures each block incorporates diverse context from the full dataset while remaining complementary to other blocks, thereby exhibiting a soft-local characteristic, balancing both global and local information.
$$\min_{\pi \in \Pi(V, \mathcal{C})} \sum_{A_i \in \pi} f(A_i) + f(\pi)$$

$f(\pi) = f(\bigcup_{A_i \in \pi} A_i)$ is the submodular function evaluation for the union of items selected in all $A_i \in \pi$.

\textbf{Local Diverse} method selects contexts from within each cluster to fully represent its internal diversity, while ensuring that the collection of blocks jointly covers the entire dataset.
$$\max_{\pi \in \Pi(V, \mathcal{C})} \sum_{A_i \in \pi} f(A_i)$$

Here $\mathcal{C}$ encodes the cluster information.

\textbf{Local Coherent} method selects context from each cluster that is closely aligned with the cluster centroid, ensuring that each block captures coherent information. Together, the blocks maintain complementary coverage of the dataset.
$$\min_{\pi \in \Pi(V, \mathcal{C})} \max_{A_i \in \pi} f(A_i)$$

\newpage
\section{Detailed Algorithm}
Below are the detailed algorithms to optimize for the objectives for \ours.
\subsection{Global Diverse}
Below are the details for the Global Diverse algorithm, which facilitates each block to summarize the whole dataset. (Alg.~\ref{alg.Global Diverse})
\begin{algorithm}[h]
\caption{Global Diverse}
\label{alg:block_Global Diverse_separate}
\begin{algorithmic}[1]
\Require Embedding matrix $E \in \mathbb{R}^{N \times d}$; number of selections $k$; block count $B$
\Ensure Partitioned selection indices $G = \{G_1, \dots, G_B\}$

\State Compute similarity matrix $S \gets E E^\top$
\State Initialize groups $G_1, \dots, G_B \gets \emptyset$
\State Initialize similarity scores $\text{sim}_b(i) \gets 0$, for all $b \in [1, B], i \in [1, N]$
\State Initialize remaining index set $R \gets \{1, \dots, N\}$; selected count $c \gets 0$

\While{$R \ne \emptyset$ and $c < k$}
    \State $b^\star \gets \arg\min_b \sum_i \text{sim}_b(i)$ \Comment{Least informative block}
    \State Compute gain for all $i \in R$: $\text{gain}(i) \gets \sum_j \max(0, S_{i,j} - \text{sim}_{b^\star}(j))$
    \State $i^\star \gets \arg\max_{i \in R} \text{gain}(i)$
    \State $G_{b^\star} \gets G_{b^\star} \cup \{i^\star\}$
    \State $\text{sim}_{b^\star}(j) \gets \max(\text{sim}_{b^\star}(j), S_{i^\star,j})$, for all $j$
    \State $R \gets R \setminus \{i^\star\}$; $c \gets c + 1$
\EndWhile

\State \Return $G$
\label{alg.Global Diverse}
\end{algorithmic}
\end{algorithm}
\newpage

\subsection{Global-Local Diverse}
Below are the Algorithm details for the Global-Local Diverse algorithm, which requires the block content to be diverse while maintaining blocks to be complementary with each other. (Alg.~\ref{Alg. Global-Local Diverse})
\begin{algorithm}[h]
\caption{Global-Local Diverse}
\label{alg:cluster_max_separate_simplified}
\begin{algorithmic}[1]
\Require Similarity Matrix $S \in \mathbb{R}^{N \times d}$, total selections $k$, number of blocks $B$
\State Selected index groups $G = \{G_1, \dots, G_B\}$

\State Initialize $G_1, \dots, G_B \gets \emptyset$; $R \gets \{1,\dots,N\}$
\State Initialize group memory $A, B \gets \emptyset$; $\text{sim}_A(i), \text{sim}_B(i) \gets 0$

\For{$b = 1$ to $B$}
    \While{$|G_b| < \lceil k/B \rceil$ and $R \ne \emptyset$}
        \State $U \gets \max(\text{sim}_A, \text{sim}_B)$
        \State Compute: $\text{gain}_U(i) + \text{gain}_B(i)$ for all $i \in R$
        \State $i^\star \gets \arg\max_{i \in R} (\text{gain}_U(i) + \text{gain}_B(i))$
        \State Add $i^\star$ to $G_b$ and $B$, update $\text{sim}_B$
        \State $R \gets R \setminus \{i^\star\}$
    \EndWhile
    \State Merge $B$ into $A$, reset $B$, update $\text{sim}_A$
\EndFor
\State \Return $G$
\label{Alg. Global-Local Diverse}
\end{algorithmic}
\end{algorithm}

\newpage

\subsection{Local Diverse}
The Local Diverse algorithm selects the most diverse contexts for each block within each cluster, preserving both intra-cluster diversity and inter-block complementarity. (Alg.~\ref{Alg. Local Diverse})

\begin{algorithm}[h]
\caption{Local Diverse}
\label{alg:cluster_max_separate}
\begin{algorithmic}[1]
\Require Embedding matrix $E \in \mathbb{R}^{N \times d}$; number of selections $k$; block count $B$
\Ensure Block assignments $G = \{G_1, \dots, G_B\}$

\State $S \gets E E^\top$ \Comment{Pairwise similarity matrix}
\State Initialize group memory: $A \gets \emptyset$, $B \gets \emptyset$
\State Initialize similarity scores $\text{sim}_A(i) \gets 0$, $\text{sim}_B(i) \gets 0$
\State $R \gets \{1, \dots, N\}$ \Comment{Remaining indices}
\State $m \gets \lceil k / B \rceil$ \Comment{Block size per group}
\State Initialize $G_1, \dots, G_B \gets \emptyset$

\For{$b = 1$ to $B$}
    \While{size of $G_b < m$ and $|R| > 0$}
        \State $U \gets \max(\text{sim}_A, \text{sim}_B)$
        \ForAll{$i \in R$}
            \State $\text{gain}_U(i) \gets \sum_j \max(0, S_{i,j} - U(j))$
            \State $\text{gain}_B(i) \gets \sum_j \max(0, S_{i,j} - \text{sim}_B(j))$
        \EndFor
        \State $i^\star \gets \arg\max_{i \in R} (\text{gain}_U(i) + \text{gain}_B(i))$
        \State $G_b \gets G_b \cup \{i^\star\}$; $B \gets B \cup \{i^\star\}$
        \State Update $\text{sim}_B \gets \textsc{BestSimilarity}(S, B)$
        \State $R \gets R \setminus \{i^\star\}$
    \EndWhile
    \State $A \gets A \cup B$; $B \gets \emptyset$
    \State Update $\text{sim}_A \gets \textsc{BestSimilarity}(S, A)$
    \State Reset $\text{sim}_B(i) \gets 0$ for all $i$
\EndFor
\State \Return $G = \{G_1, \dots, G_B\}$
\label{Alg. Local Diverse}
\end{algorithmic}
\end{algorithm}
\newpage

\subsection{Local Coherent}
Similar to Local Diverse, Local Coherent also operates at the cluster level of the entire dataset. However, it selects contexts that are similar to the cluster centroid to ensure local coherence while still maintaining inter-block complementarity. (Alg.~\ref{Alg. Local Coherent})
\begin{algorithm}[h]
\caption{Local Coherent}
\label{alg:block_minmax_separate}
\begin{algorithmic}[1]
\Require Similarity matrix $S \in \mathbb{R}^{N \times N}$; number of selections $k$; number of blocks $B$
\Ensure Partitioned index groups $G = \{G_1, \dots, G_B\}$

\State Initialize $G_1, \dots, G_B \gets \emptyset$, $R \gets \{1, \dots, N\}$, $c \gets 0$, $m \gets \lfloor k/B \rfloor$
\State Initialize $\text{sim}_\text{init} \gets 0$

\For{$b = 1$ to $B$}
    \State $i^\star \gets \arg\max_{i \in R} \sum_j \max(0, S_{i,j} - \text{sim}_\text{init}(j))$
    \State $G_b \gets \{i^\star\}$, update $\text{sim}_b \gets \max(\text{sim}_\text{init}, S_{i^\star})$
    \State $R \gets R \setminus \{i^\star\}$, $c \gets c + 1$
\EndFor

\For{$b = 1$ to $B$}
    \For{$t = 1$ to $m - 1$}
        \State Compute gain: $\text{gain}(i) = \sum_j \max(0, S_{i,j} - \text{sim}_b(j))$ for $i \in R$
        \State $i^\star \gets \arg\min_{i \in R} \text{gain}(i)$
        \State $G_b \gets G_b \cup \{i^\star\}$, update $\text{sim}_b \gets \max(\text{sim}_b, S_{i^\star})$
        \State $S[i^\star,:] \gets 0$, $R \gets R \setminus \{i^\star\}$, $c \gets c + 1$
    \EndFor
\EndFor

\State \Return $G$
\label{Alg. Local Coherent}
\end{algorithmic}
\end{algorithm}

\newpage
\section{Show Case}
Below we show some prompt examples obtained from the SST-2 dataset using our different \ours strategy. 

\subsection{Global Diverse}
\newtcbox{\promptbox}{enhanced, colframe=black!60!white,
  colback=gray!5, boxrule=0.5pt, arc=2pt, left=1mm, right=1mm, top=1mm, bottom=1mm}

\begin{tcolorbox}[title=Example Group Prompts (Sentiment Classification)]
\textbf{Group 1}
\begin{itemize}[leftmargin=1.5em, itemsep=0.1em]
  \item Sentence: all the hallmarks of a movie \textbf{(positive)}
  \item Sentence: achingly human \textbf{(positive)}
  \item Sentence: so much farcical \textbf{(negative)}
  \item Sentence: a brilliant piece of filmmaking \textbf{(positive)}
\end{itemize}

\textbf{Group 2}
\begin{itemize}[leftmargin=1.5em, itemsep=0.1em]
  \item Sentence: unlike lots of hollywood fluff \textbf{(positive)}
  \item Sentence: an ingenious and often harrowing \textbf{(positive)}
  \item Sentence: a lazy exercise in bad filmmaking \textbf{(negative)}
  \item Sentence: such good humor \textbf{(positive)}
\end{itemize}

\textbf{Group 3}
\begin{itemize}[leftmargin=1.5em, itemsep=0.1em]
  \item Sentence: has all the hallmarks of a movie \textbf{(positive)}
  \item Sentence: incredibly hokey \textbf{(negative)}
  \item Sentence: a beautifully \textbf{(positive)}
  \item Sentence: yet surprisingly entertaining \textbf{(positive)}
\end{itemize}

\textbf{Group 4}
\begin{itemize}[leftmargin=1.5em, itemsep=0.1em]
  \item Sentence: a true cinematic knack \textbf{(positive)}
  \item Sentence: so incredibly inane \textbf{(negative)}
  \item Sentence: it's not a brilliant piece of filmmaking \textbf{(negative)}
  \item Sentence: engross \textbf{(positive)}
\end{itemize}
\end{tcolorbox}
\newpage

\subsection{Global-Local Diverse}
\begin{tcolorbox}[title=Example Group Prompts (Global-Local Diverse), colback=gray!5, colframe=black!40!white]

\textbf{Group 1}
\begin{itemize}[leftmargin=1.5em, itemsep=0.2em]
  \item Sentence: all the hallmarks of a movie \textbf{(positive)}
  \item Sentence: such a stultifying \textbf{(negative)}
  \item Sentence: such good humor \textbf{(positive)}
  \item Sentence: 's not a brilliant piece of filmmaking \textbf{(positive)}
\end{itemize}

\textbf{Group 2}
\begin{itemize}[leftmargin=1.5em, itemsep=0.2em]
  \item Sentence: has all the hallmarks of a movie \textbf{(positive)}
  \item Sentence: engross \textbf{(positive)}
  \item Sentence: a lazy exercise in bad filmmaking \textbf{(negative)}
  \item Sentence: a bewilderingly brilliant and entertaining \textbf{(positive)}
\end{itemize}

\textbf{Group 3}
\begin{itemize}[leftmargin=1.5em, itemsep=0.2em]
  \item Sentence: unlike lots of hollywood fluff \textbf{(positive)}
  \item Sentence: a beautifully \textbf{(positive)}
  \item Sentence: a soulless jumble of ineptly \textbf{(negative)}
  \item Sentence: a deftly entertaining film \textbf{(positive)}
\end{itemize}

\textbf{Group 4}
\begin{itemize}[leftmargin=1.5em, itemsep=0.2em]
  \item Sentence: achingly human \textbf{(positive)}
  \item Sentence: like a film that strays past the two and a half mark \textbf{(negative)}
  \item Sentence: a remarkable piece of filmmaking \textbf{(positive)}
\end{itemize}

\end{tcolorbox}
\newpage

\subsection{Local Diverse}
\begin{tcolorbox}[title=Example Group Prompts (Local Diverse), colback=gray!5, colframe=black!40!white]

\textbf{Group 1}
\begin{itemize}[leftmargin=1.5em, itemsep=0.2em]
  \item Sentence: a well-crafted film that is all the more remarkable because it \textbf{(positive)}
  \item Sentence: a very compelling, sensitive, intelligent and almost cohesive piece of film entertainment \textbf{(positive)}
  \item Sentence: thanks to confident filmmaking and a pair of fascinating performances \textbf{(positive)}
  \item Sentence: a fiercely clever and subtle film \textbf{(positive)}
\end{itemize}

\textbf{Group 2}
\begin{itemize}[leftmargin=1.5em, itemsep=0.2em]
  \item Sentence: a tale full of nuance and character dimension \textbf{(positive)}
  \item Sentence: an ingenious and often harrowing \textbf{(positive)}
  \item Sentence: a powerful and telling story \textbf{(positive)}
  \item Sentence: an extraordinary poignancy, and the story \textbf{(positive)}
\end{itemize}

\textbf{Group 3}
\begin{itemize}[leftmargin=1.5em, itemsep=0.2em]
  \item Sentence: splendid \textbf{(positive)}
  \item Sentence: a beautifully \textbf{(positive)}
  \item Sentence: a beautifully \textbf{(positive)}
  \item Sentence: brilliant and \textbf{(positive)}
\end{itemize}

\textbf{Group 4}
\begin{itemize}[leftmargin=1.5em, itemsep=0.2em]
  \item Sentence: a great piece of filmmaking \textbf{(positive)}
  \item Sentence: a brilliant piece of filmmaking \textbf{(positive)}
  \item Sentence: more than a movie \textbf{(positive)}
  \item Sentence: a brilliant motion picture \textbf{(positive)}
\end{itemize}

\end{tcolorbox}

\newpage

\subsection{Local Coherent}
\begin{tcolorbox}[
  title=Example Group Prompts (Local Coherent), 
  colback=gray!5, 
  colframe=black!40!white,
  fontupper=\footnotesize,
  before upper=\setlength{\itemsep}{0.15em}\setlength{\leftmargini}{1.2em}
]

\textbf{Group 1}
\begin{itemize}[leftmargin=1.5em, itemsep=0.2em]
  \item Sentence: all the hallmarks of a movie \textbf{(positive)}
  \item Sentence: must be given to the water-camera operating team of Don King, Sonny Miller, and Michael Stewart. \textbf{(negative)}
  \item Sentence: David Spade as Citizen Kane? \textbf{(negative)}
  \item Sentence: Leave it to John Sayles to take on developers, the chamber of commerce, tourism, historical pageants, and commercialism all in the same movie ... without neglecting character development for even one minute. \textbf{(positive)}
\end{itemize}

\textbf{Group 2}
\begin{itemize}[leftmargin=1.5em, itemsep=0.2em]
  \item Sentence: achingly human \textbf{(positive)}
  \item Sentence: Neo-Augustinian theology: is God \textbf{(positive)}
  \item Sentence: As famous prima donna Floria Tosca, Roberto Alagna as her lover Mario Cavaradossi, and Ruggero as the villainous, lecherous police chief Scarpia, \textbf{(positive)}
  \item Sentence: Angela Gheorghiu as famous prima donna Floria Tosca, Roberto Alagna as her lover Mario Cavaradossi, and Ruggero as the villainous, lecherous police chief Scarpia, all sing beautifully and act adequately. \textbf{(positive)}
\end{itemize}

\textbf{Group 3}
\begin{itemize}[leftmargin=1.5em, itemsep=0.2em]
  \item Sentence: so much farcical \textbf{(negative)}
  \item Sentence: It’s common knowledge that Park and his founding partner, Yong Kang, lost Kozmo in the end. \textbf{(positive)}
  \item Sentence: ... built on the premise that middle-class Arkansas consists of monster truck-loving good ol' boys and peroxide blond honeys whose worldly knowledge comes from TV reruns and supermarket tabloids. \textbf{(negative)}
  \item Sentence: Couldn't someone take Rob Schneider and have him switch bodies with a funny person? \textbf{(negative)}
\end{itemize}

\textbf{Group 4}
\begin{itemize}[leftmargin=1.5em, itemsep=0.2em]
  \item Sentence: a brilliant piece of filmmaking \textbf{(positive)}
  \item Sentence: Schütte's dramatic snapshot of the artist three days before his death offers an interesting bit of speculation as to the issues Brecht faced as his life drew to a close. \textbf{(positive)}
  \item Sentence: Writer and director Otar Iosseliani's pleasant tale about a factory worker who escapes for a holiday in Venice \textbf{(positive)}
  \item Sentence: Writer and director Otar Iosseliani's pleasant tale about a factory worker who escapes for a holiday in Venice reveals how we all need a playful respite from the grind to refresh our souls. \textbf{(positive)}
\end{itemize}

\end{tcolorbox}

\end{document}